# An Evaluation of Transfer Learning for Classifying Sales Engagement Emails at Large Scale

Yong Liu
Outreach Corporation
Seattle, USA
yong.liu@outreach.io

Pavel Dmitriev
Outreach Corporation
Seattle, USA
pavel.dmitriev@outreach.io

Yifei Huang
Outreach Corporation
Seattle, USA
yifei.huang@outreach.io

Andrew Brooks
Outreach Corporation
Seattle, USA
andrew.brooks@outreach.io

Li Dong
Outreach Corporation
Seattle, USA
li.dong@outreach.io

*Abstract*—This paper conducts an empirical investigation to evaluate transfer learning for classifying sales engagement emails arising from digital sales engagement platforms. Given the complexity of content and context of sales engagement, lack of standardized large corpora and benchmarks, limited labeled examples and heterogenous context of intent, this real-world use case poses both a challenge and an opportunity for adopting a transfer learning approach. We propose an evaluation framework to assess a high performance transfer learning (HPTL) approach in three key areas in addition to commonly used accuracy metrics: 1) effective embeddings and pretrained language model usage, 2) minimum labeled samples requirement and 3) transfer learning implementation strategies. We use in-house sales engagement email samples as the experiment dataset, which includes over 3000 emails labeled as positive, objection, unsubscribe, or not-sure. We discuss our findings on evaluating BERT, ELMo, Flair and GloVe embeddings with both feature-based and fine-tuning approaches and their scalability on a GPU cluster with increasingly larger labeled samples. Our results show that fine-tuning of the BERT model outperforms with as few as 300 labeled samples, but underperforms with fewer than 300 labeled samples, relative to all the feature-based approaches using different embeddings.

Keywords— *pretrained language model, BERT, ELMo, Flair embeddings, GloVe, transfer learning, email intent classification, sales engagement*

I. INTRODUCTION

The recent success of pretrained language models (LMs) such as BERT [22] topping the leaderboards of natural language understanding benchmarks has garnered increasing interest in the general machine learning community for adopting transfer learning in natural language processing. However, whether such benchmark successes can be readily translated to well performing models in the real-world applications and use cases is still an open question.

In this paper, we evaluate and assess the most recently released state-of-the-art pretrained LMs and embeddings for their efficacy as a foundation to a high performance transfer learning (HPTL) application for a real use case: email intent classification in the sales engagement domain. A transfer learning is usually considered high performance if it meets satisfactory accuracy metric with fewer training examples and quicker train/test time than traditional methods, although there is currently no universally applicable evaluation framework yet for natural language processing applications [6].

The email research community has long been relying on some publicly available email datasets such as Enron and Avocado etc. [12]. However, unlike other text classification scenarios, the email domain is filled with unique challenges depending on the nature of communication. This is especially true for the recently risen digital sales engagement platforms where emails became one of the large scale modern tools that sales professionals use.

Sales is one of the oldest professions on earth. Until very recently, a sales representative (rep) typically got a list of names (leads, or prospects) and manually went through the list one by one calling and emailing them. The rise of Sales Engagement Platforms (SEPs) such as Outreach.io, SalesLoft.com, InsideSales.com, Groove.co and Apollo.io is rapidly transforming the sales process. SEP encodes a company's sales process as a sequence of steps consisting of emails, phone calls, LinkedIn messages, and other tasks. Different sequences are used for different types of prospects, market segments, etc. SEP then completely automates execution of some sales tasks (e.g. auto-sending personalized emails and LinkedIn messages), while scheduling and reminding the rep when it is the right time to do the manual tasks (e.g. phone call, custom manual email). As a result, every rep can simultaneously perform one-on-one personalized outreach to up to 10x more prospects than before. SEPs also give rise to a new set of challenges to achieve better automation and next action recommendation based on the understanding of email intent.

Unlike the general domain of text understanding, where large corpora are readily available, there is no standardized sales engagement email corpus now and in foreseeable future due to GDPR (General Data Protection Regulation) and other privacy and enterprise security concerns. Thus the approaches such as those proposed by [18][15] which begin with a pretraining of a domain-relevant large corpus are not viable at this moment.

This leads us to investigate augmenting existing pretrained LMs from other domains, with a small amount of in-domain training data prepared in-house. In this paper we report on a number of experiments we performed along with an evaluation framework proposed for measuring performance of a HPTL model. We attempt to answer the following key research questions:1) What kind of embeddings work best? 2) What kind of transfer learning implementation strategies work best? and 3) How much labeled data is enough?

While this paper attempts to answer these questions for the email classification task in the sales engagement domain, we believe our findings can be leveraged in other domains and bring practitioners a step closer to understanding the advantages and limitations of HPTL in practical applications.

The paper is organized as follows: Section II discusses some motivating use cases and unique technical challenges in the sales engagement domain; Section III presents the evaluation framework and experiment design; Section IV describes the results and our discussions; Section V presents some related work and Section VI presents concluding remarks.

## II. USE CASES AND TECHNICAL CHALLENGES OF LARGE-SCALE SALES ENGAGEMENT EMAILS

In this section, we aim to introduce some selected motivating use cases and technical challenges in the sales engagement domain.

### A. Use Cases

Marketing and sales professionals invest substantial resources in sales content. One important type of content in SEP is outbound sales engagement email templates (i.e., sent by sales reps to prospects), which typically include self-introduction of salesperson, value propositions of product, and a call to action (e.g., asking for time for a phone call or meeting).

The status quo metrics for measuring performance of email templates are email open rate, link click rate and reply rate. However, this approach does not consider the fact that not all replies from prospects are equal. For example, an overly aggressive email usually results in higher reply rate, compared to a more polite one, but most of those replies would be of negative tone or unsubscription requests. Even coarse-grain classification of intent of prospects' replies into *Positive, Objection, Unsubscription, or Not-Sure* categories is very useful and allows for much better evaluation of template performance. More fine-grain classification which identifies the type of objections raised by the prospect can provide deeper and more actionable insights.

SEPs also improves the efficiency of the sales process at scale - by optimizing the effectiveness of the commonly used automated tasks. Techniques such as A/B testing can be used to test different variations of email templates, and pick the best one. Performance measurement based on intent classification enables more effective optimization through A/B testing – the intent based metrics are much better at identifying the winning email template than the reply rate metric.

### B. Technical Challenges

***Heterogeneous Context and Players***: SEP is the communication platform for the entire B2B sales lifecycle, including sales development phase (e.g., cold contacting to confirm buying interests), deal closing phase (e.g., demonstrate product value, pass security review and negotiate contract), customer success phase (e.g., onboarding, contract renewal and expansion). In modern B2B sales, especially in tech industry, each phase has its own specialized department and job roles. Consequently, the nature of email communication highly depends on its context and players. This leads to significant challenges for email intent classification for SEP. Not only is there a generalization problem across different customers a single SEP serves (e.g., small family businesses to multinational corporations), the heterogeneity in context leads to generalization problems even within the same customer as the sales process progresses. We have observed that the positive reply rate increases and the unsubscription reply rate decreases as the sales process progresses further. Thus a model that's trained only on the initial replies will not perform well on subsequent corresponding emails. Email threading structure further increases the context complexities.

***Limited Labeled Sales Engagement Domain Emails:*** Obtaining quality labels cannot be easily outsourced to contractors due to the sensitivity of business emails. It generally has to be done in house under a high level of supervision and security, which is time-consuming and labor/cost intensive. For an SEP to serve thousands of client companies, enabling customers to propose new intents uniquely relevant for their business is important. Thus, it is desirable to learn new intents from small samples (e.g., less than a few hundred examples).

## III. EVALUATION FRAMEWORK AND EXPERIMENT DESIGN

We propose an evaluation framework and a set of experiments to answer our research questions in section I. In this paper, we focus on addressing the second challenge mentioned in Section II where the number of labeled examples is limited.

### A. Evaluation Framework

While traditional supervised machine learning models are trained and evaluated on datasets from the same domain, it is commonly acknowledged that performance will degrade when applied to a different context or domain. Thus, our evaluation framework for an HPTL approach is based on the following considerations:

*1) Different pretrained language models and embeddings:* Character-based, string-based and word-based pretrained embeddings and language models are considered either alone or stacked together.

*2) Minimum number of labeled examples needed for achieving satisfactory accuracy:* We need to find out how quickly the accuracy improves when using pretrained LMs as the size of the labeled training dataset increases.

*3) Fine-tuning vs. feature-based transfer learning implementation strategies:* Feature-based approaches use pretrained representations as input features for the downstream task without touching the original models, while fine-tuning approaches train the downstream task model by fine-tuning pretrained model parameters. We consider both approaches and evaluate how quickly they reach satisfactory performance.

### B. Pretrained Language Models and Embeedings

In this paper, we select the following representational pretrained LMs and classical embeddings based on both their different embedding properties and their state-of-the-art performance reported in the literature on public benchmark datasets.

*1) GloVe[7]:* this is one of the classical word embeddings. This is included here for comparison and for stacking (combined with other pretrained LMs) with feature-based approaches.

*2) ELMo[2]:* this is one of the context-aware character-based embeddings that is based on a recurrent neural network architecture.

*3) Flair[16]:* this is a contextual string embedding that reported good performance for text classification. We use both

the news-forward and news-backward pretrained flair embeddings.

*4) BERT[22]:* this is the state-of-the-art transformer-based language model released in late 2018 by Google. BERT uses a variant of Byte-Pair Encoding called WordPieces to tokenize the text (e.g.: "calling" can be tokenized to ["call", "##ing"])). We use the bert-base-uncased pretrained language model in this paper.

MT-DNN built on top of BERT from Microsoft [3] and GPT-2 from OpenAI [4] both recently reported new state-of-the-art pretrained language models results but their code and models are yet to be released, thus they are not included in this evaluation. Sentence-level embeddings have shown minimum benefit of gaining better accuracy [14][23], and thus they are not included in this evaluation.

*C. Experiments Dataset and Runs*

We use an in-house labeled email dataset, which consists of 3840 initial replies from prospects to sales development representatives. Tables I describes the labels and the number of emails under each label. We randomize these emails and split them into 80/10/10 for training, validation and testing.

TABLE I.  EXPERIMENT EMAIL DATASET LABELS AND SAMPLE COUNTS

| Email Label | Total Count | Train Sample Count | Validation Sample Count | Test Sample Count |
|---|---|---|---|---|
| notsure | 209 | 171 | 18 | 20 |
| objection | 2218 | 1789 | 213 | 216 |
| positive | 761 | 600 | 75 | 86 |
| unsubscribe | 652 | 512 | 78 | 62 |

We then run the following experiments:

*1) Feature-based transfer learning experiments:* First, encode each email using one or multiple stacked embeddings at the sentence-level representation, which serves as the initial input to a document-level (i.e., email-level) embedding layer before the classification layer. In this paper, two different document-level embedding layers are compared. The first one is a pooling model which computes the sentence-level mean of a multi-sentence email text embeddings. The second one is a randomly initialized single-layer 512-dimension LSTM (Long Short-Term Memory) neural network with a reprojected 256-dimension output embedding vector. The classification layer uses a linear layer (pytorch's `torch.nn.linear`) and a softmax to predict the final class label. A document-level pooling layer is non-parametric and computationally efficient, which is why it is chosen as a possible alternative to a more advanced LSTM layer, which is usually recommended [22]. The Flair framework (`https://github.com/zalandoresearch/flair`) is used to run these experiments. Each experiment was run with 20 epochs. Since using BERT in a feature-based approach requires a separate pooling to obtain a token level embedding from subword embedding due to its WordPiece tokenization, two different BERT token-level pooling are used: one is to use the first subword embedding only(denoted as '`bert`'), the other is to use the mean of all subword embeddings(denoted as '`bert_mean`' in the rest of the paper).The last four layers of the pretrained BERT transformer were used, as recommended in [22].

*2) Fine-tuning experiments using BERT:* Since BERT is the latest state-of-the-art pretrained LM that achieves the top benchmark score, we evaluate BERT using the fine-tuning approach (denoted as '`bert_finetuning`').We use the original Google implementation of BERT on Tensorflow. A few key parameters for fine-tuning BERT are as follows: `MAX_SEQ_LENGTH` is set to be 128, `BATCH_SIZE` is 32, `NUM_TRAIN_EPOCHS` is 3.0. and `LEARNING_RATE` is 2e-5.

*3) Effect of labeled training size on performance:* We run experiments using the top-performing embeddings for the feature-based approach and fine-tuning BERT approach, with increasing numbers of labeled training examples: 50, 100, 200, 300, 500, 1000, 2000, 3000 to evaluate how quickly the performance of the transfer learning models improves.

Note that there is no hyperparameter sweeping experiments in this paper, which could have improved the accuracy further.

All experiments were run on an AWS (Amazon Web Service) p2.xlarge GPU (Graphics Process Unit) cluster with 61 GB memory hosted by Databricks.com, a cloud-based unified analytics platform for big data and high performance machine learning workloads.

TABLE II.  RESULTS OF DIFFERENT PRETRAINED LMS AND EMBEDDINGS

| Pretrained LMs and Embeddings | Configurations and Performance Metrics | | | |
|---|---|---|---|---|
|  | Micro Average f1-score (mean±std) | | Elapsed Time (mins)(mean±std) | |
|  | Pooling | LSTM | Pooling | LSTM |
| bert | 0.769±0.032 | 0.808±0.013 | 3.0±1.0 | 5.3±1.1 |
| bert+elmo | 0.759±0.091 | 0.814±0.004 | 4.9±1.8 | 7.3±2.1 |
| bert+flair | 0.791±0.017 | 0.806±0.006 | 3.4±1.4 | 6.2±1.6 |
| bert+glove | 0.772±0.013 | 0.798±0.008 | 0.9±0.4 | 3.5±0.4 |
| bert_mean | 0.786±0.032 | 0.820±0.012 | 2.8±0.9 | 5.3±1.1 |
| bert_mean+elmo | 0.786±0.018 | 0.807±0.013 | 4.6±1.6 | 7.1±2.3 |
| bert_mean+flair | 0.790±0.025 | 0.817±0.013 | 3.5±1.4 | 6.2±1.7 |
| bert_mean+glove | 0.784±0.011 | 0.811±0.025 | 4.1±0.8 | 7.2±1.6 |
| elmo | 0.747±0.026 | 0.791±0.006 | 2.0±0.8 | 4.9±0.9 |
| elmo+flair | 0.773±0.038 | 0.796±0.015 | 2.5±1.2 | 5.5±1.3 |
| elmo+glove | 0.783±0.005 | 0.824±0.014 | 1.0±0.4 | 3.4±0.4 |
| flair | 0.783±0.003 | 0.791±0.013 | 1.0±0.4 | 4.0±0.4 |
| flair+glove | 0.797±0.018 | 0.799±0.017 | 2.5±0.4 | 5.2±0.5 |
| glove | 0.751±0.011 | **0.831±0.013** | 2.0±0.0 | 4.0±0.2 |
| bert_finetuning | **0.874±0.005** | | 8.6±0.0 | |

## IV. RESULTS AND DISCUSSIONS

### A. Results of Using Different Pretrained Language Models and Embeddings

For each pretrained LMs and embeddings and their combinations in Table II, we run five training/testing runs with the exact same configuration and parameters. The mean and standard deviation of micro-average f1-score and elapsed time in minutes of these five runs are listed in the table. All elapsed time excludes the time to download the pretrained language models from the internet and downloading time is typically under 45 seconds . All runs in this set of experiments use the full set of the training data. Except for the last row in Table II, all runs use the feature-based approach. That last row in Table II is the fine-tuning BERT experiment, thus there is no separate document pooling or LSTM layer.

For the feature-based approach, GloVe embeddings alone work slightly better than all other recently released embeddings and pretrained LMs. This indicates that for this task, context free

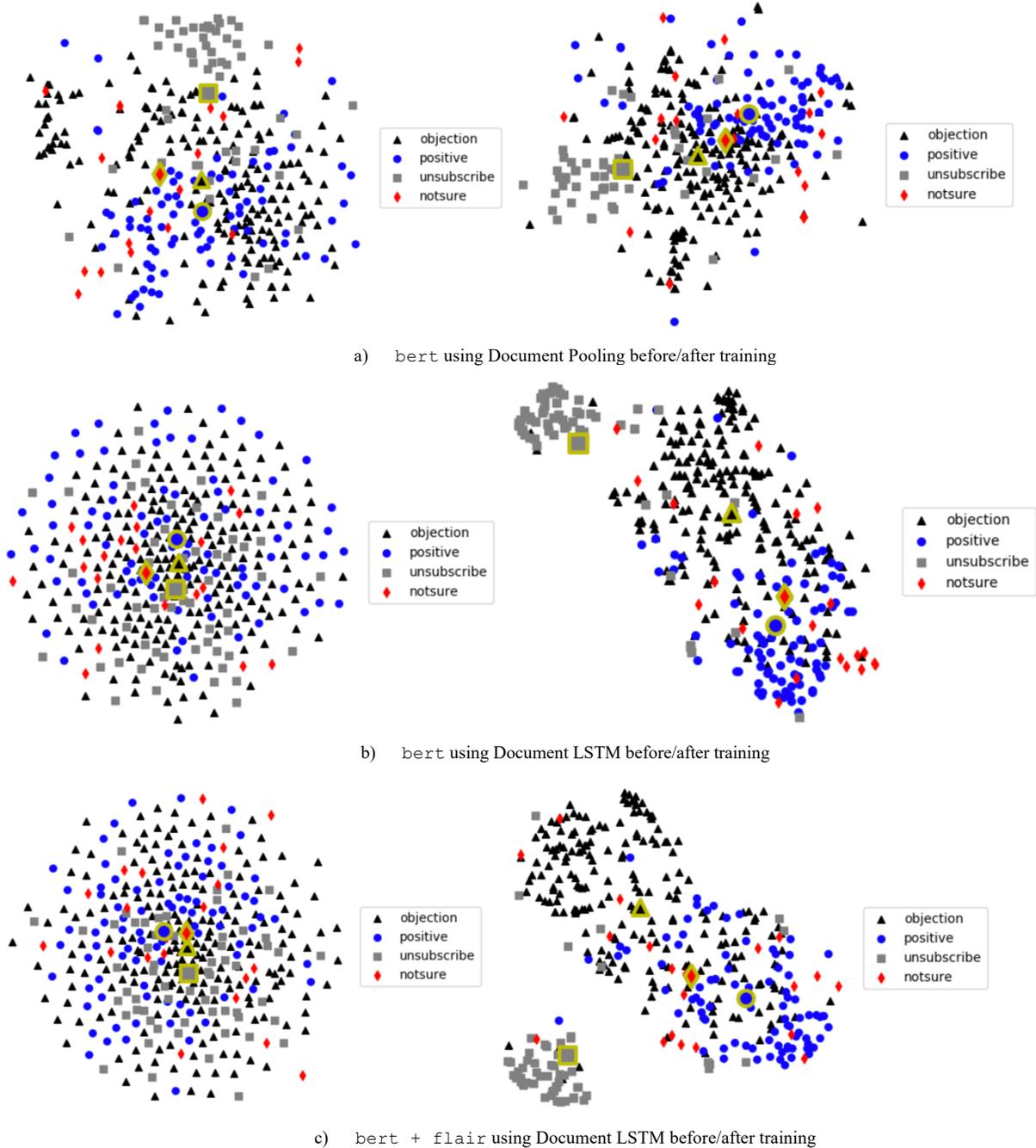

a) `bert` using Document Pooling before/after training

b) `bert` using Document LSTM before/after training

c) `bert + flair` using Document LSTM before/after training

Fig. 1. tSNE Visualization of Email Test Dataset Cluster Change Before and After Training for Selected Feature-based Approaches

word-level feature representation might be sufficient to most of the signal to differentiate the classes.

We also notice that performance using a document pooling layer is consistently lower than the document LSTM layer. This could be explained by the fact that averaging a multi-sentence email embedding to the sentence-level could lose some of the key semantic knowledge initially gained from the pretrained language model. Particularly if specific words or n-grams alone have a pronounced influence on the signal, this could be diluted by averaging.

To illustrate the behavior of both the choice of embeddings/pretrained LMs and document-level layer, a selected set of tSNE (t-distributed Stochastic Neighbor Embedding) visualizations [25] are shown in Fig.1. The four different colors and shapes represent the four different labels and the dots represent examples from the test datasets. For each row of plots, the left plot shows the test dataset cluster at the initialization stage, and the right plot shows the cluster at the end of training. If a cluster is around the centroid of the same color-coded and shaped label, then that means the model has a good separation mechanism. Note that using `bert` alone with a pooling layer has a very good initialization before training, but the cluster becomes less separable after training. On the other hand, using `bert` alone plus an LSTM layer has a randomized initialization, but has a better final separable cluster. For `bert+flair` with a document LSTM combination, the final cluster is much cleanly separable. In addition, we notice that "not-sure" labeled examples are present in all other clusters, indicating that reliably separating them out is challenging.

Fine-tuning BERT outperforms all feature-based approaches with a large margin and an almost constant elapsed time of 8.6 minutes.

*B. Results for Labeled Training Size with Feature-based and Fine-tuning Approach*

We ran a set of experiments to evaluate the performance on how a pretrained language model-based classifier scale with the number of labeled training examples. We use the top performing models showed in Table II, which includes the best overall performing model (fine-tuning BERT) and the four top feature-based models (`glove`, `elmo+glove`, `bert_mean`, and `bert_mean+flair`). The input training examples are randomly sampled to produce an increasingly larger datasets for training. The hold-out test dataset for all experiments remains the same. For each training size, five runs were conducted and the average and standard deviation of the micro-average f1-score were computed and plotted. Fig. 2 shows the results for a training sample size range described in section III.

It seems that at around 2000 samples we are reaching a micro-average f1-score of 0.8 for the top three feature-based approaches, while `elmo+glove` only has a score of 0.765. This gives us a good estimate of the minimum number of labeled samples required to achieve a good performance for using feature-based approach. It is interesting to see that even though `elmo+glove` is ranked the second best (below `glove`) when the full training set is used (see Table II), its f1-score is almost consistently lower than all other feature-based approaches when train size is small, except at around 500 samples where `glove` has the lowest mean f1-score and a large standard deviation. It is also noticeable that stacking `bert_mean+flair` has a very small mixed effect in terms of f1-score, compared with using `bert_mean` alone. When the sample size is 50, `bert_mean+flair` has the best f1-score (0.6693), slightly better than `bert_mean` (0.6578), but then these two cross-over slightly multiple times when the train size grows. The performance of `glove` zig-zags a bit, compared with other feature-based approaches, exhibited by a large standard deviation when the training sample size is less than or equal to 500, which implies the unstable performance of using GloVe with smaller training datasets.

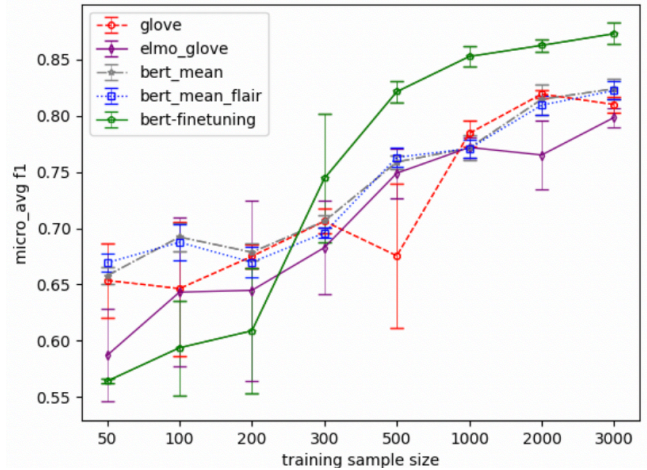

Fig.2. Scaling Effect of Micro_avg_f1-score vs. Labeled Sample Size

For the fine-tuning approach using BERT, the f1-score initially is lower than that of the feature-based approach, but then quickly surpasses the latter at around 300 labeled samples. It then reaches a score of 0.8214 with 500 labeled samples before plateauing at 0.8729 with 3000 samples. The initial lower scores may indicate that a pretrained LM such as BERT may have memorized some knowledge that's not transferable to this new task and it needs some warm-up training sets to adapt. However, once it passes the "warm-up" stage, fine-tuning a pretrained LM does take fewer labeled samples to get good and stable performance with very small standard deviation. Also, much higher standard deviation with the f1-score is noticeable when the labeled training size is around a few hundred, a potentially noisy stage during the fine-tuning process.

*C. Discussions*

Our experiments provide some key takeaways and answers to our research questions listed in Section I.

**Combination of string and WordPiece embeddings works slightly better for feature-based transfer learning when training sample size is small, while Glove works slightly better when all train samples are used**: For example, with as little as 50 labeled examples, the f1-score of `bert_mean+flair` reaches 0.6693, the best among all feature-based approaches and even fine-tuning BERT. It seems that the input representations using BERT and Flair combine the best of both string-based and WordPiece encoding of the underlying texts. Such a stacking embedding approach is a good strategy (see also [4]) for a feature-based transferring. However, this advantage is not sustainable when the training size increases. Using Glove alone works slightly better among all feature-based approaches when all training samples are used.

*A document-level representation layer with average pooling has a decent initialization, but its performance is surpassed by a LSTM document layer representation after training*: Specifically, average pooling of a multi-sentence email embedding may provide a good initialization, but its initial lead quickly disappears. On the other hand, representing an email with an LSTM document layer can have a randomized initialization but then rapidly outperforms the average pooling representation. It seems that using more intelligent pooling such as fuzzy set [20] may gain better performance in the end. This may also suggest that we need different architectures for different classification tasks [5][8].

*Achieving satisfactory performance still requires a moderate amount of training data*: The size of available labeled training examples is a constraint in our study. Recent literature talks about zero-shot or few-shot learning [9][13][21] for text classification, which is exciting. The results in this paper show that indeed the pretrained LMs solve the cold start problem when there is very little training data. However, to get to a satisfactory performance, it may still need two thousand examples for a feature-based approach or 500 examples for fine-tuning a pretrained BERT language model.

*Fine-tuning a pretrained LM substantially improves performance*: For fine-tuning a pretrained LM such as BERT, it does show great potential to transfer the pretrained linguistics knowledge with as little as 300 samples and can continue to improve with more samples before reaching a plateau after 1000 samples. This seems to be a promising direction for building a HPTL model. In this paper, we only explore the standard fine-tuning approach [22], which potentially requires adjusting millions of pretrained parameters. A more efficient approach such as [19] may be used to progressively freeze and unfreeze the layers for fine tuning, which may be explored in our future study.

## V. Related Work

Yogatama et al. [6] evaluated the general linguistics intelligence and transferred knowledge for BERT and ELMo and proposed a new metric called codelength to measure "the ability to generalize rapidly to a new task by using previously acquired knowledge.". They found that both models (BERT and ELMo) were able to approach their asymptotic errors after seeing approximately 40,000 training examples, a surprisingly high number of examples for models that rely on pretrained models. They did not evaluate the combination of different pretrained language models and embeddings, but their findings on the training size requirement are revealing.

Radford et al. [4] reported how to evaluate "generalization and memorization" of a pretrained language model by proposing a method using a blooming filter to measure the 8-gram overlap between training and testing datasets. They also reported the importance of input representation: string-based, word-based or Byte-Pair Encoding and combining these would yield better performance. This is consistent with our experiment results although they did not evaluate BERT or Flair or their combinations.

Zhai et al. [1] conducted intrinsic evaluation on GloVe where they found word embedding clusters give high correlations to the synonym and hyponym sets in WordNet, and extrinsic evaluation on using GloVe for two NLP tasks: named entity recognition and part-of-speech tagging. They did not evaluate the effect of training size or fine-tuning approach for text classification.

An additional transfer learning implementation strategy called "adaptor-based" [11] was recently presented as a third way (in addition to feature-based and fine-tuning based) to do transfer learning in a very parameter efficient way. This might open doors for continuous learning where different tasks can be fine-tuned only on a particular set of adaptor related parameters. However, they haven't released their source code and pretrained model as of this writing, thus we did not include it in this study.

## VI. Conclusions and Future Work

In this paper, we presented selected use cases and unique technique challenges posed by the SEPs and conducted a set of experiments along with an evaluation framework for assessing an HPTL model that is based on pretrained LMs and embeddings. Our experiments show that fine-tuning a pretrained language model such as BERT for classifying sales engagement emails with a 4-class labels and moderate class imbalance requires as few as 500 training examples to reach a micro-average f1-score of 0.82, which could be sufficient in this domain. We also find that a feature-based approach using pretrained language models and embeddings performs less well but may provide a better starting point if the number of labeled examples is less than 300. The combination of BERT and Flair embeddings in a feature-based approach shows that encoding of the input text does make a slight difference when the training example size is small, while GloVe alone works slightly better when all training samples are used. An HPTL system should make a trade-off decision based on the availability of labeled data.

We are just scratching the surface of evaluating and moving towards developing an HPTL for the sales engagement domain. Our future work will look into feasibility of building HPTL across multiple diverse SEP customers. We will also look into how we can add new labels with multi-task learning[17], joint embeddings of both samples and labels [10] and continuous learning[11], which will further unlock the power of transfer learning. In addition, how we can leverage unlabeled data using transductive transfer learning [24] and combine it with the power of pretrained LMs to do few-shot learning in the sales engagement domain is of great interest to us.

Finally, we hope that our work inspires more evaluations of HPTL in real world scenarios, helping practitioners better understand capabilities and limitations of these techniques and determine the best configurations for their scenarios.

## Acknowledgment

The authors thank Emi Ayada who contributed to an earlier investigation, and Erin Miller and Chris Heisey who annotated the email dataset.